\documentclass[11pt]{article}
\usepackage[latin9]{inputenc}
\usepackage{array}
\usepackage{verbatim}
\usepackage{multirow}
\usepackage{amsmath}
\usepackage{graphicx}
\usepackage[unicode=true]
 {hyperref}
\usepackage{breakurl}

\makeatletter

\providecommand{\tabularnewline}{\\}

\usepackage{enumitem}		

\@ifundefined{date}{}{\date{}}
\usepackage{acl2016a}
\usepackage{times}
\usepackage{latexsym}

\aclfinalcopy 
\def\aclpaperid{337} 

\title{Generative Topic Embedding: a Continuous Representation of Documents (Extended Version with Proofs)}

\author{Shaohua Li$^{1,2}$ \hspace{3em} Tat-Seng Chua$^1$ \quad\quad\quad Jun Zhu$^3$ \quad\quad\quad\quad\quad Chunyan Miao$^2$ \\
    shaohua@gmail.com \quad dcscts@nus.edu.sg \quad dcszj@tsinghua.edu.cn \quad ascymiao@ntu.edu.sg \\
         1. School of Computing, National University of Singapore \\  
         2. Joint NTU-UBC Research Centre of Excellence in Active Living for the Elderly (LILY) \\     
         3. Department of Computer Science and Technology, Tsinghua University }

\usepackage{amsmath,amssymb}
\usepackage{scrextend}
\usepackage{footmisc}
\usepackage[titletoc,title]{appendix}

\usepackage{tikz}
\usetikzlibrary{arrows} 
\usetikzlibrary{shapes}
\usetikzlibrary{fit}
\usetikzlibrary{chains}
\usetikzlibrary{calc}
\usetikzlibrary{bayesnet}
\usetikzlibrary{decorations.markings}

\DeclareMathOperator*{\T}{\scriptscriptstyle \top}
\newcommand{\script}{\scriptscriptstyle}

\newcommand{\vast}{\bBigg@{3}}
\newcommand{\Vast}{\bBigg@{4}}

\makeatother

\begin{document}
\maketitle
\begin{abstract}
Word embedding maps words into a low-dimensional continuous embedding
space by exploiting the local word collocation patterns in a small
context window. On the other hand, topic modeling maps documents onto
a low-dimensional topic space, by utilizing the global word collocation
patterns in the same document. These two types of patterns are complementary.
In this paper, we propose a generative topic embedding model to combine
the two types of patterns. In our model, topics are represented by
embedding vectors, and are shared across documents. The probability
of each word is influenced by both its local context and its topic.
A variational inference method yields the topic embeddings as well
as the topic mixing proportions for each document. Jointly they represent
the document in a low-dimensional continuous space. In two document
classification tasks, our method performs better than eight existing
methods, with fewer features. In addition, we illustrate with an example
that our method can generate coherent topics even based on only one
document.
\end{abstract}

\section{Introduction}

Representing documents as fixed-length feature vectors is important
for many document processing algorithms. Traditionally documents are
represented as a bag-of-words (BOW) vectors. However, this simple
representation suffers from being high-dimensional and highly sparse,
and loses semantic relatedness across the vector dimensions.

Word Embedding methods have been demonstrated to be an effective way
to represent words as continuous vectors in a low-dimensional embedding
space \cite{bengio,word2vec,glove,levy}. The learned embedding for
a word encodes its semantic/syntactic relatedness with other words,
by utilizing local word collocation patterns. In each method, one
core component is the \emph{embedding link function}, which predicts
a word's distribution given its context words, parameterized by their
embeddings.

When it comes to documents, we wish to find a method to encode their
overall semantics. Given the embeddings of each word in a document,
we can imagine the document as a ``bag-of-vectors''. Related words
in the document point in similar directions, forming \emph{semantic
clusters}. The centroid of a semantic cluster corresponds to the most
representative embedding of this cluster of words, referred to as
the \emph{semantic centroids}. We could use these semantic centroids
and the number of words around them to represent a document.

In addition, for a set of documents in a particular domain, some semantic
clusters may appear in many documents. By learning collocation patterns
across the documents, the derived semantic centroids could be more
topical and less noisy.

Topic Models, represented by Latent Dirichlet Allocation (LDA) \cite{lda},
are able to group words into topics according to their collocation
patterns across documents. When the corpus is large enough, such patterns
reflect their semantic relatedness, hence topic models can discover
coherent topics. The probability of a word is governed by its latent
topic, which is modeled as a categorical distribution in LDA. Typically,
only a small number of topics are present in each document, and only
a small number of words have high probability in each topic. This
intuition motivated \newcite{lda} to regularize the topic distributions
with Dirichlet priors.

Semantic centroids have the same nature as topics in LDA, except that
the former exist in the embedding space. This similarity drives us
to seek the common semantic centroids with a model similar to LDA.
We extend a generative word embedding model PSDVec \cite{psdvec},
by incorporating topics into it. The new model is named TopicVec.
In TopicVec, an embedding link function models the word distribution
in a topic, in place of the categorical distribution in LDA. The advantage
of the link function is that the semantic relatedness is already encoded
as the \emph{cosine} distance in the embedding space. Similar to LDA,
we regularize the topic distributions with Dirichlet priors. A variational
inference algorithm is derived. The learning process derives \emph{topic
embeddings} in the same embedding space of words. These topic embeddings
aim to approximate the underlying semantic centroids.

To evaluate how well TopicVec represents documents, we performed two
document classification tasks against eight existing topic modeling
or document representation methods. Two setups of TopicVec outperformed
all other methods on two tasks, respectively, with fewer features.
In addition, we demonstrate that TopicVec can derive coherent topics
based only on \emph{one }document, which is not possible for topic
models.

The source code of our implementation is available at \href{https://github.com/askerlee/topicvec}{https://github.com/askerlee/topicvec}.

\section{Related Work}

\newcite{psdvec} proposed a generative word embedding method PSDVec,
which is the precursor of TopicVec. PSDVec assumes that the conditional
distribution of a word given its context words can be factorized approximately
into independent log-bilinear terms. In addition, the word embeddings
and regression residuals are regularized by Gaussian priors, reducing
their chance of overfitting. The model inference is approached by
an efficient Eigendecomposition and blockwise-regression method \cite{psdvec-osp}.
TopicVec differs from PSDVec in that in the conditional distribution
of a word, it is not only influenced by its context words, but also
by a topic, which is an embedding vector indexed by a latent variable
drawn from a Dirichlet-Multinomial distribution.

\newcite{replicated-softmax} proposed to model topics as a certain
number of binary hidden variables, which interact with all words in
the document through weighted connections. \newcite{docNADE} assigned
each word a unique topic vector, which is a summarization of the context
of the current word.

\newcite{huang-global} proposed to incorporate global (document-level)
semantic information to help the learning of word embeddings. The
global embedding is simply a weighted average of the embeddings of
words in the document.

\newcite{doc2vec} proposed Paragraph Vector. It assumes each piece
of text has a latent paragraph vector, which influences the distributions
of all words in this text, in the same way as a latent word. It can
be viewed as a special case of TopicVec, with the topic number set
to 1. Typically, however, a document consists of multiple semantic
centroids, and the limitation of only one topic may lead to underfitting.

\newcite{lftm} proposed Latent Feature Topic Modeling (LFTM), which
extends LDA to incorporate word embeddings as latent features. The
topic is modeled as a mixture of the conventional categorical distribution
and an embedding link function. The coupling between these two components
makes the inference difficult. They designed a Gibbs sampler for model
inference. Their implementation\footnote{https://github.com/datquocnguyen/LFTM/}
is slow and infeasible when applied to a large corpous.

\newcite{liu} proposed Topical Word Embedding (TWE), which combines
word embedding with LDA in a simple and effective way. They train
word embeddings and a topic model separately on the same corpus, and
then average the embeddings of words in the same topic to get the
embedding of this topic. The topic embedding is concatenated with
the word embedding to form the topical word embedding of a word. In
the end, the topical word embeddings of all words in a document are
averaged to be the embedding of the document. This method performs
well on our two classification tasks. Weaknesses of TWE include: 1)
the way to combine the results of word embedding and LDA lacks statistical
foundations; 2) the LDA module requires a large corpus to derive semantically
coherent topics.

\newcite{gaussianLDA} proposed Gaussian LDA. It uses pre-trained
word embeddings. It assumes that words in a topic are random samples
from a multivariate Gaussian distribution with the topic embedding
as the mean. Hence the probability that a word belongs to a topic
is determined by the Euclidean distance between the word embedding
and the topic embedding. This assumption might be improper as the
Euclidean distance is not an optimal measure of semantic relatedness
between two embeddings\footnote{Almost all modern word embedding methods adopt the exponentiated cosine
similarity as the link function, hence the cosine similarity may be
assumed to be a better estimate of the semantic relatedness between
embeddings derived from these methods.}.

\section{Notations and Definitions}

Throughout this paper, we use uppercase bold letters such as $\boldsymbol{S},\boldsymbol{V}$
to denote a matrix or set, lowercase bold letters such as $\boldsymbol{v}_{w_{i}}$
to denote a vector, a normal uppercase letter such as $N,W$ to denote
a scalar constant, and a normal lowercase letter as $s_{i},w_{i}$
to denote a scalar variable.

Table \ref{tab:topicvec-Notation-Table} lists the notations in this
paper.

In a document, a sequence of words is referred to as a \textit{text
window}, denoted by $w_{i},\cdots,w_{i+l}$, or $w_{i}{:}w_{i+l}$.
A text window of chosen size $c$ before a word $w_{i}$ defines the
\textit{context} of $w_{i}$ as $w_{i-c},\cdots,w_{i-1}$. Here $w_{i}$
is referred to as the \textit{focus word}. Each context word $w_{i-j}$
and the focus word $w_{i}$ comprise a\textit{ }\textit{\emph{bigram}}
$w_{i-j},w_{i}$.

\begin{table}
\centering{}%
\begin{tabular}{cc}
\hline 
Name  & Description\tabularnewline
\hline 
{\footnotesize{}$\boldsymbol{S}$}  & {\footnotesize{}Vocabulary $\{s_{1},\cdots,s_{\script W}\}$}\tabularnewline
{\footnotesize{}$\boldsymbol{V}$}  & \textit{\emph{\footnotesize{}Embedding matrix $(\boldsymbol{v}_{s_{1}},\cdots,\boldsymbol{v}_{s_{\script W}})$}}\tabularnewline
{\footnotesize{}$\boldsymbol{D}$}  & {\footnotesize{}Document set $\{d_{1},\cdots,d_{\script M}\}$}\tabularnewline
{\footnotesize{}$\boldsymbol{v}_{s_{i}}$}  & \textit{\emph{\footnotesize{}Embedding }}{\footnotesize{}of word $s_{i}$}\tabularnewline
{\footnotesize{}$a_{s_{i}s_{j}},\boldsymbol{A}$}  & \textit{\emph{\footnotesize{}Bigram residuals}}\tabularnewline
$\boldsymbol{t}_{ik},\boldsymbol{T}_{i}$ & {\footnotesize{}Topic embeddings in doc $d_{i}$}\tabularnewline
$r_{ik},\boldsymbol{r}_{i}$ & {\footnotesize{}Topic residuals in doc $d_{i}$}\tabularnewline
$z_{ij}$ & {\footnotesize{}Topic assignment of the $j$-th word $j$ in doc $d_{i}$}\tabularnewline
$\boldsymbol{\phi}_{i}$ & {\footnotesize{}Mixing proportions of topics in doc $d_{i}$}\tabularnewline
\hline 
\end{tabular}\caption{\label{tab:topicvec-Notation-Table}Table of notations}
\end{table}

We assume each word in a document is semantically similar to a \textit{topic
embedding}. Topic embeddings reside in the same $N$-dimensional space
as word embeddings. When it is clear from context, topic embeddings
are often referred to as \textit{topic}s. Each document has $K$ candidate
topics, arranged in the matrix form $\boldsymbol{T}_{i}=(\boldsymbol{t}_{i1}\cdots\boldsymbol{t}_{iK})$,
referred to as the \textit{topic matrix}. Specifically, we fix $\boldsymbol{t}_{i1}=\boldsymbol{0}$,
referring to it as the \emph{null topic}.

In a document $d_{i}$, each word $w_{ij}$ is assigned to a topic
indexed by $z_{ij}\in\{1,\cdots,K\}$. Geometrically this means the
embedding $\boldsymbol{v}_{w_{ij}}$ tends to align with the direction
of $\boldsymbol{t}_{i,z_{ij}}$. Each topic $\boldsymbol{t}_{ik}$
has a document-specific prior probability to be assigned to a word,
denoted as $\phi_{ik}=P(k|d_{i})$. The vector $\boldsymbol{\phi}_{i}=(\phi_{i1},\cdots,\phi_{iK})$
is referred to as the \textit{mixing proportions }of these topics
in document $d_{i}$.

\section{Link Function of Topic Embedding}

In this section, we formulate the distribution of a word given its
context words and topic, in the form of a link function.

The core of most word embedding methods is a\emph{ link function}
that connects the embeddings of a focus word and its context words,
to define the distribution of the focus word. \newcite{psdvec} proposed
the following link function:\vspace{-0.25in}

\begin{align}
 & P(w_{c}\mid w_{0}:w_{c-1})\nonumber \\
\approx & P(w_{c})\exp\biggl\{\boldsymbol{v}_{w_{c}}^{\T}\sum_{l=0}^{c-1}\boldsymbol{v}_{w_{l}}+\sum_{l=0}^{c-1}a_{w_{l}w_{c}}\biggr\}.\label{eq:groupcond}
\end{align}

Here $a_{w_{l}w_{c}}$ is referred as the bigram residual, indicating
the non-linear part not captured by $\boldsymbol{v}_{w_{c}}^{\T}\boldsymbol{v}_{w_{l}}$.
It is essentially the logarithm of the normalizing constant of a softmax
term. Some literature, e.g. \cite{glove}, refers to such a term as
a bias term.

\eqref{eq:groupcond} is based on the assumption that the conditional
distribution $P(w_{c}\mid w_{0}:w_{c-1})$ can be factorized approximately
into independent log-bilinear terms, each corresponding to a context
word. This approximation leads to an efficient and effective word
embedding algorithm \emph{PSDVec} \cite{psdvec}. We follow this assumption,
and propose to incorporate the topic of $w_{c}$ in a way like a latent
word. In particular, in addition to the context words, the corresponding
embedding $\boldsymbol{t}_{ik}$ is included as a new log-bilinear
term that influences the distribution of $w_{c}$. Hence we obtain
the following extended link function:\vspace{-0.25in}\begin{addmargin}[-0.5em]{0em}

\begin{align}
 & \:P(w_{c}\mid w_{0}{:}w_{c-1},z_{c},d_{i})\approx P(w_{c})\cdot\nonumber \\
 & \exp\Bigl\{\boldsymbol{v}_{w_{c}}^{\T}\Bigl(\sum_{l=0}^{c-1}\boldsymbol{v}_{w_{l}}+\boldsymbol{t}_{z_{c}}\Bigr)\negmedspace+\negmedspace\sum_{l=0}^{c-1}a_{w_{l}w_{c}}\negmedspace+\negmedspace r_{z_{c}}\Bigr\},\label{eq:conddistCT}
\end{align}
\end{addmargin}where $d_{i}$ is the current document, and $r_{z_{c}}$
is the logarithm of the normalizing constant, named the \emph{topic
residual}. Note that the topic embeddings $\boldsymbol{t}_{z_{c}}$
may be specific to $d_{i}$. For simplicity of notation, we drop the
document index in $\boldsymbol{t}_{z_{c}}$. To restrict the impact
of topics and avoid overfitting, we constrain the magnitudes of all
topic embeddings, so that they are always within a hyperball of radius
$\gamma$.

It is infeasible to compute the exact value of the topic residual
$r_{k}$. We approximate it by the context size $c=0$. Then \eqref{eq:conddistCT}
becomes: 
\begin{equation}
P(w_{c}\mid k,d_{i})=P(w_{c})\exp\left\{ \boldsymbol{v}_{w_{c}}^{\T}\boldsymbol{t}_{k}+r_{k}\right\} .\label{eq:conddistT}
\end{equation}

It is required that $\sum_{w_{c}\in\boldsymbol{S}}P(w_{c}\mid k)=1$
to make \eqref{eq:conddistT} a distribution. It follows that 
\begin{equation}
r_{k}=-\log\Bigl(\sum_{s_{j}\in\boldsymbol{S}}P(s_{j})\exp\{\boldsymbol{v}_{s_{j}}^{\T}\boldsymbol{t}_{k}\}\Bigr).\label{eq:r_ik}
\end{equation}

\eqref{eq:r_ik} can be expressed in the matrix form:
\begin{equation}
\boldsymbol{r}=-\log(\boldsymbol{u}\exp\{\boldsymbol{V}^{\T}\boldsymbol{T}\}),\label{eq:r_mat}
\end{equation}
where $\boldsymbol{u}$ is the row vector of unigram probabilities.

\begin{addmargin}[-1em]{0em}
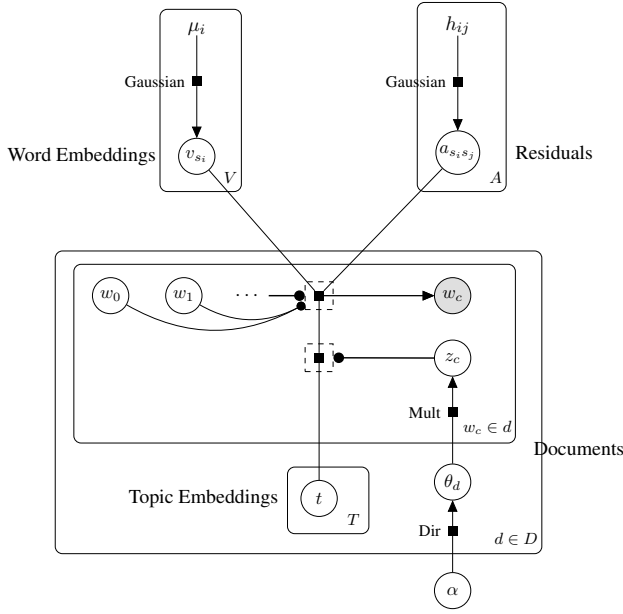
\begin{figure}[t]
\centering{}\scalebox{0.68}{\usetikzlibrary{bayesnet} \begin{tikzpicture}[x=2.2cm,y=2cm]  \node[latent]    (w1)      {$w_1$} ;   \node[draw=none, right=of w1, xshift=-0.5cm] (dots) {$\cdots$} ;  \node[latent, left=of w1, xshift=1.5cm]    (w0)      {$w_0$} ;   \node[obs, right=of dots, xshift=1cm]  (wc)  {$w_c$} ;   \node[latent, below=of wc, yshift=1.5cm]    (zc)      {$z_c$} ;   \node[latent, below=of zc, yshift=0.3cm]    (theta-d)      {$\theta_d$} ;   \node[latent, below=of theta-d, yshift=0.6cm]    (alpha)      {$\alpha$} ;     \node[latent, above=of w1, xshift=0.25cm] (vsi) {$v_{s_i}$} ;  \node[const, above=of vsi] (mu_i) {$\mu_i$} ;  \node[left=of vsi, xshift=0.7cm] (vsi_desc) {Word Embeddings};
 \node[latent, above=of wc, xshift=0.1cm] (aij) {$a_{s_i s_j}$} ;  \node[const, above=of aij, yshift=-0.15cm] (hij) {$h_{ij}$} ;  \node[right=of aij, xshift=-0.45cm] (aij_desc) {Residuals};
 \factor[below=of mu_i] {vsi-prior} {left:Gaussian} {} {} ;  \factor[below=of hij] {aij-prior} {left:Gaussian} {} {} ;  \factor[below=of zc, yshift=0.2cm] {zc-dist} {left:Mult} {} {} ;  \factor[below=of theta-d, yshift=0.3cm] {theta-d-prior} {left:Dir} {} {} ; \tikzset{every label/.style={yshift=-1.3cm, xshift=-0.5cm}}  \factor[right=of dots, dashed] {addv} {} {} {} ;  \factor[below=of addv, yshift=-0.22cm, dashed] {topic} {} {} {} ;
\node[latent, below=of addv, yshift=-1.5cm]    (t)      {$t$} ; 
\node[left=of t, xshift=0.7cm] (t_desc) {Topic Embeddings};
\tikzset{rectangle/.append style={inner sep=5pt}}  \gate {addv-gate} {(addv)} {dots}  \draw [-*] (w1) to[out=-30,in=210] (addv-gate);  \draw [-*] (w0) to[out=-30,in=210] (addv-gate);  \factoredge {mu_i} {vsi-prior} {vsi} ;  \factoredge {hij} {aij-prior} {aij} ;  \factoredge {vsi} {addv} {wc} ;  \factoredge {aij} {addv} {wc} ;  \factoredge {theta-d} {zc-dist} {zc} ;  \factoredge {alpha} {theta-d-prior} {theta-d} ;
 \gate {topic-gate} {(topic)} {zc} \edge[-] {t} {topic} ; \edge[-] {topic} {addv} ;
\tikzstyle{plate} = [draw, rectangle, rounded corners, fit=#1, inner sep=7pt] \tikzstyle{plate caption} = [caption, node distance=0cm, inner sep=2pt, outer sep=-4pt, below left=4pt and -20pt of #1.south east]
 \plate [xshift=-0.1cm] {plate1}{ (w0)(w1)(addv)(wc)(zc)(zc-dist) } {$w_c \in d$}
\tikzstyle{plate caption} = [caption, node distance=0cm, inner sep=2pt, outer sep=-4pt, below left=0pt and -10pt of #1.south east]
 \plate {plateT}{ (t) } { $T$ }
 \plate [xshift=-0.1cm] {plate2}{ (plate1)(theta-d)(theta-d-prior)(plateT) } {$d \in D$}
 \node[right=of plate2, xshift=-1.3cm, yshift=-0.9cm] {Documents};
 \plate [xshift=-0.1cm] {}{ (mu_i) (vsi-prior) (vsi) } {$V$}  \plate [xshift=-0.1cm] {}{ (hij) (aij) } {$A$}
\end{tikzpicture} }\caption{\label{fig:topic-embedding}Graphical representation of TopicVec.}
\end{figure}
\end{addmargin}

\section{Generative Process and Likelihood}

The generative process of words in documents can be regarded as a
hybrid of LDA and PSDVec. Analogous to PSDVec, the word embedding
$\boldsymbol{v}_{s_{i}}$ and residual $a_{s_{i}s_{j}}$ are drawn
from respective Gaussians. For the sake of clarity, we ignore their
generation steps, and focus on the topic embeddings. The remaining
generative process is as follows:
\begin{enumerate}[leftmargin=1em,topsep=4pt,itemsep=-0.6ex]
\item For the $k$-th topic, draw a topic embedding uniformly from a hyperball
of radius $\gamma$, i.e. $\boldsymbol{t}_{k}\sim\textrm{Unif}(B_{\gamma})$;
\item For each document $d_{i}$:

\begin{enumerate}[leftmargin=1em,topsep=0pt,itemsep=-0.6ex]
\item Draw the mixing proportions $\boldsymbol{\phi}_{i}$ from the Dirichlet
prior $\textrm{Dir}(\boldsymbol{\alpha})$;
\item For the $j$-th word: 

\begin{enumerate}[leftmargin=1em]
\item Draw topic assignment $z_{ij}$ from the categorical distribution
Cat$(\boldsymbol{\phi}_{i})$;
\item Draw word $w_{ij}$ from $\boldsymbol{S}$ according to $P(w_{ij}\mid w_{i,j-c}{:}w_{i,j-1},z_{ij},d_{i})$.
\end{enumerate}
\end{enumerate}
\end{enumerate}
The above generative process is presented in plate notation in Figure
\eqref{fig:topic-embedding}.

\subsection{Likelihood Function}

Given the embeddings $\boldsymbol{V}$, the bigram residuals $\boldsymbol{A}$,
the topics $\boldsymbol{T}_{i}$ and the hyperparameter $\boldsymbol{\alpha}$,
the complete-data likelihood of a single document $d_{i}$ is:
\begin{align}
 & p(d_{i},\boldsymbol{Z}_{i},\boldsymbol{\phi}_{i}|\boldsymbol{\alpha},\boldsymbol{V},\boldsymbol{A},\boldsymbol{T}_{i})\nonumber \\
= & p(\boldsymbol{\phi}_{i}|\boldsymbol{\alpha})p(\boldsymbol{Z}_{i}|\boldsymbol{\phi}_{i})p(d_{i}|\boldsymbol{V},\boldsymbol{A},\boldsymbol{T}_{i},\boldsymbol{Z}_{i})\nonumber \\
= & \frac{\Gamma(\sum_{k=1}^{K}\alpha_{k})}{\prod_{k=1}^{K}\Gamma(\alpha_{k})}\prod_{j=1}^{K}\phi_{ij}^{\alpha_{j}-1}\cdot\prod_{j=1}^{L_{i}}\Bigg(\phi_{i,z_{ij}}P(w_{ij})\nonumber \\
 & \cdot\exp\biggl\{\boldsymbol{v}_{w_{ij}}^{\T}\Bigl(\sum_{l=j-c}^{j-1}\boldsymbol{v}_{w_{il}}+\boldsymbol{t}_{z_{ij}}\Bigr)\nonumber \\
 & +\negthickspace\sum_{l=j-c}^{j-1}a_{w_{il}w_{ij}}\negmedspace+r_{i,z_{ij}}\biggr\}\Bigg),\label{eq:jointdoc}
\end{align}
where $\boldsymbol{Z}_{i}=(z_{i1},\cdots,z_{iL_{i}})$, and $\Gamma(\cdot)$
is the Gamma function.

\vspace{8pt}
Let $\boldsymbol{Z},\boldsymbol{T},\boldsymbol{\phi}$ denote the
collection of all the document-specific $\{\boldsymbol{Z}_{i}\}_{i=1}^{M},\{\boldsymbol{T}_{i}\}_{i=1}^{M},\{\boldsymbol{\phi}_{i}\}_{i=1}^{M}$,
respectively. Then the complete-data likelihood of the whole corpus
is:
\begin{align}
 & p(\boldsymbol{D},\boldsymbol{A},\boldsymbol{V},\boldsymbol{Z},\boldsymbol{T},\boldsymbol{\phi}|\boldsymbol{\alpha},\gamma,\boldsymbol{\mu})\nonumber \\
= & \prod_{i=1}^{W}P(\boldsymbol{v}_{s_{i}};\mu_{i})\prod_{i,j=1}^{W,W}P(a_{s_{i}s_{j}};f(h_{ij}))\prod_{k}^{K}\textrm{Unif}(B_{\gamma})\nonumber \\
 & \cdot\prod_{i=1}^{M}\left\{ p(\boldsymbol{\phi}_{i}|\boldsymbol{\alpha})p(\boldsymbol{Z}_{i}|\boldsymbol{\phi}_{i})p(d_{i}|\boldsymbol{V},\boldsymbol{A},\boldsymbol{T}_{i},\boldsymbol{Z}_{i})\right\} \nonumber \\
= & \frac{1}{\mathcal{Z}(\boldsymbol{H},\boldsymbol{\mu})U_{\gamma}^{K}}\exp\{-\negmedspace\sum_{i,j=1}^{W,W}f(h_{i,j})a_{s_{i}s_{j}}^{2}\negmedspace-\negmedspace\sum_{i=1}^{W}\mu_{i}\Vert\boldsymbol{v}_{s_{i}}\Vert^{2}\}\nonumber \\
 & \cdot\prod_{i=1}^{M}\biggl\{\frac{\Gamma(\sum_{k=1}^{K}\alpha_{k})}{\prod_{k=1}^{K}\Gamma(\alpha_{k})}\prod_{j=1}^{K}\phi_{ij}^{\alpha_{j}-1}\cdot\prod_{j=1}^{L_{i}}\biggl(\phi_{i,z_{ij}}P(w_{ij})\nonumber \\
 & \cdot\exp\Bigl\{\boldsymbol{v}_{w_{ij}}^{\T}\Bigl(\sum_{l=j-c}^{j-1}\boldsymbol{v}_{w_{il}}\negmedspace+\negmedspace\boldsymbol{t}_{z_{ij}}\Bigr)\negmedspace+\negmedspace\negthickspace\sum_{l=j-c}^{j-1}\negmedspace a_{w_{il}w_{ij}}\negmedspace+\negmedspace r_{i,z_{ij}}\negmedspace\Bigr\}\negmedspace\biggr)\negmedspace\biggr\},\label{eq:jointcorpus}
\end{align}
where $P(\boldsymbol{v}_{s_{i}};\mu_{i})$ and $P(a_{s_{i}s_{j}};f(h_{ij}))$
are the two Gaussian priors as defined in \cite{psdvec}. Following
the convention in \cite{psdvec}, $h_{ij},\boldsymbol{H}$ are empirical
bigram probabilities, $\boldsymbol{\mu}$ are the embedding magnitude
penalty coefficients, and $\mathcal{Z}(\boldsymbol{H},\boldsymbol{\mu})$
is the normalizing constant for word embeddings. $U_{\gamma}$ is
the volume of the hyperball of radius $\gamma$. 

Taking the logarithm of both sides, we obtain 
\begin{align}
 & \log p(\boldsymbol{D},\boldsymbol{A},\boldsymbol{V},\boldsymbol{Z},\boldsymbol{T},\boldsymbol{\phi}|\boldsymbol{\alpha},\gamma,\boldsymbol{\mu})\nonumber \\
= & C_{0}-\log\mathcal{Z}(\boldsymbol{H},\boldsymbol{\mu})-\Vert\boldsymbol{A}\Vert_{f(\boldsymbol{H})}^{2}-\sum_{i=1}^{W}\mu_{i}\Vert\boldsymbol{v}_{s_{i}}\Vert^{2}\nonumber \\
+ & \sum_{i=1}^{M}\biggl\{\sum_{k=1}^{K}\log\phi_{ik}(m_{ik}+\alpha_{k}-1)+\sum_{j=1}^{L_{i}}\biggl(r_{i,z_{ij}}\nonumber \\
+ & \boldsymbol{v}_{w_{ij}}^{\T}\Bigl(\sum_{l=j-c}^{j-1}\boldsymbol{v}_{w_{il}}+\boldsymbol{t}_{z_{ij}}\Bigr)+\negthickspace\sum_{l=j-c}^{j-1}\negthickspace a_{w_{il}w_{ij}}\biggr)\negthickspace\biggr\},\label{eq:logjointcorpusT}
\end{align}
where $m_{ik}=\sum_{j=1}^{L_{i}}\delta(z_{ij}=k)$ counts the number
of words assigned with the $k$-th topic in $d_{i}$, $C_{0}=M\log\frac{\Gamma(\sum_{k=1}^{K}\alpha_{k})}{\prod_{k=1}^{K}\Gamma(\alpha_{k})}+\sum_{i,j=1}^{M,L_{i}}\log P(w_{ij})-K\log U_{\gamma}$
is constant given the hyperparameters.

\section{Variational Inference Algorithm}

\subsection{Learning Objective and Process}

Given the hyperparameters $\boldsymbol{\alpha},\gamma,\boldsymbol{\mu}$,
the learning objective is to find the embeddings $\boldsymbol{V}$,
the topics $\boldsymbol{T}$, and the word-topic and document-topic
distributions $p(\boldsymbol{Z}_{i},\boldsymbol{\phi}_{i}|d_{i},\boldsymbol{A},\boldsymbol{V},\boldsymbol{T})$.
Here the hyperparameters $\boldsymbol{\alpha},\gamma,\boldsymbol{\mu}$
are kept constant, and we make them implicit in the distribution notations.

However, the coupling between $\boldsymbol{A},\boldsymbol{V}$ and
$\boldsymbol{T},\boldsymbol{Z},\boldsymbol{\phi}$ makes it inefficient
to optimize them simultaneously. To get around this difficulty, we
learn word embeddings and topic embeddings separately. Specifically,
the learning process is divided into two stages:
\begin{enumerate}[leftmargin=1em,topsep=4pt,itemsep=-0.6ex]
\item In the first stage, considering that the topics have a relatively
small impact to word distributions and the impact might be ``averaged
out'' across different documents, we simplify the model by ignoring
topics temporarily. Then the model falls back to the original PSDVec.
The optimal solution $\boldsymbol{V}^{*},\boldsymbol{A}^{*}$ is obtained
accordingly;
\item In the second stage, we treat $\boldsymbol{V}^{*},\boldsymbol{A}^{*}$
as constant, plug it into the likelihood function, and find the corresponding
optimal $\boldsymbol{T}^{*},p(\boldsymbol{Z},\boldsymbol{\phi}|\boldsymbol{D},\boldsymbol{A}^{*},\boldsymbol{V}^{*},\boldsymbol{T}^{*})$
of the full model. As in LDA, this posterior is analytically intractable,
and we use a simpler variational distribution $q(\boldsymbol{Z},\boldsymbol{\phi})$
to approximate it.
\end{enumerate}

\subsection{Mean-Field Approximation and Variational GEM Algorithm}

In this stage, we fix $\boldsymbol{V}=\boldsymbol{V}^{*},\boldsymbol{A}=\boldsymbol{A}^{*}$,
and seek the optimal $\boldsymbol{T}^{*},p(\boldsymbol{Z},\boldsymbol{\phi}|\boldsymbol{D},\boldsymbol{A}^{*},\boldsymbol{V}^{*},\boldsymbol{T}^{*})$.
As $\boldsymbol{V}^{*},\boldsymbol{A}^{*}$ are constant, we also
make them implicit in the following expressions.

For an arbitrary variational distribution $q(\boldsymbol{Z},\boldsymbol{\phi})$,
the following equalities hold\vspace{-4bp}
\begin{align}
 & E_{q}\log\left[\frac{p(\boldsymbol{D},\boldsymbol{Z},\boldsymbol{\phi}|\boldsymbol{T})}{q(\boldsymbol{Z},\boldsymbol{\phi})}\right]\nonumber \\
= & E_{q}\left[\log p(\boldsymbol{D},\boldsymbol{Z},\boldsymbol{\phi}|\boldsymbol{T})\right]+\mathcal{H}(q)\nonumber \\
= & \log p(\boldsymbol{D}|\boldsymbol{T})-\textrm{KL}(q||p),
\end{align}
where $p=p(\boldsymbol{Z},\boldsymbol{\phi}|\boldsymbol{D},\boldsymbol{T})$,
$\mathcal{H}(q)$ is the entropy of $q$. This implies\vspace{-4bp}
\begin{align}
 & \textrm{KL}(q||p)\nonumber \\
= & \log p(\boldsymbol{D}|\boldsymbol{T})-\Bigl(E_{q}\left[\log p(\boldsymbol{D},\boldsymbol{Z},\boldsymbol{\phi}|\boldsymbol{T})\right]+\mathcal{H}(q)\Bigr)\nonumber \\
= & \log p(\boldsymbol{D}|\boldsymbol{T})-\mathcal{L}(q,\boldsymbol{T}).\label{eq:kleqn}
\end{align}

In \eqref{eq:kleqn}, $E_{q}\left[\log p(\boldsymbol{D},\boldsymbol{Z},\boldsymbol{\phi}|\boldsymbol{T})\right]+\mathcal{H}(q)$
is usually referred to as the \textit{variational free energy} $\mathcal{L}(q,\boldsymbol{T})$,
which is a lower bound of $\log p(\boldsymbol{D}|\boldsymbol{T})$.
Directly maximizing $\log p(\boldsymbol{D}|\boldsymbol{T})$ w.r.t.
$\boldsymbol{T}$ is intractable due to the hidden variables $\boldsymbol{Z},\boldsymbol{\phi}$,
so we maximize its lower bound $\mathcal{L}(q,\boldsymbol{T})$ instead.
We adopt a mean-field approximation of the true posterior as the variational
distribution, and use a variational algorithm to find $q^{*},\boldsymbol{T}^{*}$
maximizing $\mathcal{L}(q,\boldsymbol{T})$. 

The following variational distribution is used:\vspace{-3bp}
\begin{align}
 & q(\boldsymbol{Z},\boldsymbol{\phi};\boldsymbol{\pi},\boldsymbol{\theta})=q(\boldsymbol{\phi};\boldsymbol{\theta})q(\boldsymbol{Z};\boldsymbol{\pi})\nonumber \\
= & \prod_{i=1}^{M}\left\{ \textrm{Dir}(\boldsymbol{\phi}_{i};\boldsymbol{\theta}_{i})\prod_{j=1}^{L_{i}}\textrm{Cat}(z_{ij};\boldsymbol{\pi}_{ij})\right\} .\label{eq:vardist}
\end{align}

We can obtain (Appendix \ref{sec:Derivation-of-L})\vspace{-3bp}
\begin{align}
 & \mathcal{L}(q,\boldsymbol{T})\nonumber \\
= & \sum_{i=1}^{M}\Biggl\{\sum_{k=1}^{K}\Bigl(\sum_{j=1}^{L_{i}}\pi_{ij}^{k}+\alpha_{k}-1\Bigr)\Bigl(\psi(\theta_{ik})-\psi(\theta_{i0})\Bigr)\nonumber \\
 & +\textrm{Tr}(\boldsymbol{T}_{i}^{\T}\sum_{j=1}^{L_{i}}\boldsymbol{v}_{w_{ij}}\boldsymbol{\pi}_{ij}^{\T})+\boldsymbol{r}_{i}^{\T}\sum_{j=1}^{L_{i}}\boldsymbol{\pi}_{ij}\Biggr\}\nonumber \\
 & +\mathcal{H}(q)+C_{1},\label{eq:emobj}
\end{align}
where $\boldsymbol{T}_{i}$ is the topic matrix of the $i$-th document,
and $\boldsymbol{r}_{i}$ is the vector constructed by concatenating
all the topic residuals $r_{ik}$. $C_{1}=C_{0}-\log\mathcal{Z}(\boldsymbol{H},\boldsymbol{\mu})-\Vert\boldsymbol{A}\Vert_{f(\boldsymbol{H})}^{2}-\sum_{i=1}^{W}\mu_{i}\Vert\boldsymbol{v}_{s_{i}}\Vert^{2}+\sum_{i,j=1}^{M,L_{i}}\Bigl(\boldsymbol{v}_{w_{ij}}^{\T}\sum_{k=j-c}^{j-1}\boldsymbol{v}_{w_{ik}}+\sum_{k=j-c}^{j-1}a_{w_{ik}w_{ij}}\Bigr)$
is constant.

We proceed to optimize \eqref{eq:emobj} with a Generalized Expectation-Maximization
(GEM) algorithm w.r.t. $q$ and $\boldsymbol{T}$ as follows:
\begin{enumerate}[leftmargin=1em,topsep=4pt,itemsep=-0.6ex]
\item Initialize all the topics $\boldsymbol{T}_{i}=\boldsymbol{0}$, and
correspondingly their residuals $\boldsymbol{r}_{i}=\boldsymbol{0}$;
\item Iterate over the following two steps until convergence. In the $l$-th
step:

\begin{enumerate}[leftmargin=1em,topsep=-3pt,itemsep=0ex]
\item Let the topics and residuals be $\boldsymbol{T}=\boldsymbol{T}^{(l-1)},\boldsymbol{r}=\boldsymbol{r}^{(l-1)}$,
find $q^{(l)}(\boldsymbol{Z},\boldsymbol{\phi})$ that maximizes $\mathcal{L}(q,\boldsymbol{T}^{(l-1)})$.
This is the Expectation step (E-step). In this step, $\log p(\boldsymbol{D}|\boldsymbol{T})$
is constant. Then the $q$ that maximizes $\mathcal{L}(q,\boldsymbol{T}^{(l)})$
will minimize $\textrm{KL}(q||p)$, i.e. such a $q$ is the closest
variational distribution to $p$ measured by KL-divergence;
\item Given the variational distribution $q^{(l)}(\boldsymbol{Z},\boldsymbol{\phi})$,
find $\boldsymbol{T}^{(l)},\boldsymbol{r}^{(l)}$ that improve $\mathcal{L}(q^{(l)},\boldsymbol{T})$,
using Gradient descent method. This is the generalized Maximization
step (M-step). In this step, $\boldsymbol{\pi},\boldsymbol{\theta},\mathcal{H}(q)$
are constant.
\end{enumerate}
\end{enumerate}

\subsubsection{Update Equations of $\boldsymbol{\pi},\boldsymbol{\theta}$  in E-Step}

In the E-step, $\boldsymbol{T}=\boldsymbol{T}^{(l-1)},\boldsymbol{r}=\boldsymbol{r}^{(l-1)}$
are constant. Taking the derivative of $\mathcal{L}(q,\boldsymbol{T}^{(l-1)})$
w.r.t. $\pi_{ij}^{k}$ and $\theta_{ik}$, respectively, we can obtain
the optimal solutions (Appendix \ref{sec:E-step}) at:
\begin{align}
\pi_{ij}^{k} & \propto\exp\{\psi(\theta_{ik})+\boldsymbol{v}_{w_{ij}}^{\T}\boldsymbol{t}_{ik}+r_{ik}\}.\label{eq:solPi}\\
\theta_{ik} & =\sum_{j=1}^{L_{i}}\pi_{ij}^{k}+\alpha_{k}.\label{eq:solTheta}
\end{align}

\subsubsection{Update Equation of $\boldsymbol{T}_{i}$ in M-Step}

In the Generalized M-step, $\boldsymbol{\pi}=\boldsymbol{\pi}^{(l)},\boldsymbol{\theta}=\boldsymbol{\theta}^{(l)}$
are constant. For notational simplicity, we drop their superscripts
$(l)$. 

To update $\boldsymbol{T}_{i}$, we first take the derivative of \eqref{eq:emobj}
w.r.t. $\boldsymbol{T}_{i}$, and then take the Gradient Descent method.

The derivative is obtained as (Appendix \ref{sec:dLdT}):
\begin{align}
 & \frac{\partial\mathcal{L}(q^{(l)},\boldsymbol{T})}{\partial\boldsymbol{T}_{i}}\nonumber \\
= & \sum_{j=1}^{L_{i}}\boldsymbol{v}_{w_{ij}}\boldsymbol{\pi}_{ij}^{\T}+\sum_{k=1}^{K}\bar{m}_{ik}\frac{\partial r_{ik}}{\partial\boldsymbol{T}_{i}},\label{eq:dLdT}
\end{align}
where $\bar{m}_{ik}=\sum_{j=1}^{L_{i}}\pi_{ij}^{k}=E[m_{ik}]$, the
sum of the variational probabilities of each word being assigned to
the $k$-th topic in the $i$-th document. $\frac{\partial r_{ik}}{\partial\boldsymbol{T}_{i}}$
is a gradient matrix, whose $j$-th column is $\frac{\partial r_{ik}}{\partial\boldsymbol{t}_{ij}}$. 

Remind that $r_{ik}=-\log\Bigl(E_{P(s)}[\exp\{\boldsymbol{v}_{s}^{\T}\boldsymbol{t}_{ik}\}]\Bigr)$.
When $j\ne k$, it is easy to verify that $\frac{\partial r_{ik}}{\partial\boldsymbol{t}_{ij}}=\boldsymbol{0}$.
When $j=k$, we have 
\begin{align}
\frac{\partial r_{ik}}{\partial\boldsymbol{t}_{ik}} & =e^{-r_{ik}}\cdot E_{P(s)}[\exp\{\boldsymbol{v}_{s}^{\T}\boldsymbol{t}_{ik}\}\boldsymbol{v}_{s}]\nonumber \\
 & =e^{-r_{ik}}\cdot\sum_{s\in W}\exp\{\boldsymbol{v}_{s}^{\T}\boldsymbol{t}_{ik}\}P(s)\boldsymbol{v}_{s}\nonumber \\
 & =e^{-r_{ik}}\cdot\exp\{\boldsymbol{t}_{ik}^{\T}\boldsymbol{V}\}(\boldsymbol{u}\circ\boldsymbol{V}),\label{eq:drdt}
\end{align}
where $\boldsymbol{u}\circ\boldsymbol{V}$ is to multiply each column
of $\boldsymbol{V}$ with $\boldsymbol{u}$ element-by-element.

Therefore $\frac{\partial r_{ik}}{\partial\boldsymbol{T}_{i}}=(\boldsymbol{0},\cdots\frac{\partial r_{ik}}{\partial\boldsymbol{t}_{ik}},\cdots,\boldsymbol{0})$.
Plugging it into \eqref{eq:dLdT}, we obtain
\[
\frac{\partial\mathcal{L}(q^{(l)},\boldsymbol{T})}{\partial\boldsymbol{T}_{i}}=\sum_{j=1}^{L_{i}}\boldsymbol{v}_{w_{ij}}\boldsymbol{\pi}_{ij}^{\T}+(\bar{m}_{i1}\frac{\partial r_{i1}}{\partial\boldsymbol{t}_{i1}},\cdots,\bar{m}_{iK}\frac{\partial r_{iK}}{\partial\boldsymbol{t}_{iK}}).
\]

We proceed to optimize $\boldsymbol{T}_{i}$ with a gradient descent
method:{\setlength{\belowdisplayskip}{4pt}\vspace{-6bp}
\[
\boldsymbol{T}_{i}^{(l)}=\boldsymbol{T}^{(l-1)}+\lambda(l,L_{i})\frac{\partial\mathcal{L}(q^{(l)},\boldsymbol{T})}{\partial\boldsymbol{T}_{i}},
\]
}where $\lambda(l,L_{i})=\frac{L_{0}\lambda_{0}}{l\cdot\max\{L_{i},L_{0}\}}$
is the learning rate function, $L_{0}$ is a pre-specified document
length threshold, and $\lambda_{0}$ is the initial learning rate.
As the magnitude of $\frac{\partial\mathcal{L}(q^{(l)},\boldsymbol{T})}{\partial\boldsymbol{T}_{i}}$
is approximately proportional to the document length $L_{i}$, to
avoid the step size becoming too big a on a long document, if $L_{i}>L_{0}$,
we normalize it by $L_{i}$.

To satisfy the constraint that $\|\boldsymbol{t}_{ik}^{(l)}\|\le\gamma$,
when $\boldsymbol{t}_{ik}^{(l)}>\gamma$, we normalize it by $\gamma/\|\boldsymbol{t}_{ik}^{(l)}\|$.

After we obtain the new $\boldsymbol{T}$, we update $\boldsymbol{r}_{i}^{(m)}$
using \eqref{eq:r_mat}.

Sometimes, especially in the initial several iterations, due to the
excessively big step size of the gradient descent, $\mathcal{L}(q,\boldsymbol{T})$
may decrease after the update of $\boldsymbol{T}$. Nonetheless the
general direction of $\mathcal{L}(q,\boldsymbol{T})$ is increasing.

\subsection{Sharing of Topics across Documents}

In principle we could use one set of topics across the whole corpus,
or choose different topics for different subsets of documents. One
could choose a way to best utilize cross-document information.

For instance, when the document category information is available,
we could make the documents in each category share their respective
set of topics, so that $M$ categories correspond to $M$ sets of
topics. In the learning algorithm, only the update of $\pi_{ij}^{k}$
needs to be changed to cater for this situation: when the $k$-th
topic is relevant to the document $i$, we update $\pi_{ij}^{k}$
using \eqref{eq:solPi}; otherwise $\pi_{ij}^{k}=0$.

An identifiability problem may arise when we split topic embeddings
according to document subsets. In different topic groups, some highly
similar redundant topics may be learned. If we project documents into
the topic space, portions of documents in the same topic in different
documents may be projected onto different dimensions of the topic
space, and similar documents may eventually be projected into very
different topic proportion vectors. In this situation, directly using
the projected topic proportion vectors could cause problems in unsupervised
tasks such as clustering. A simple solution to this problem would
be to compute the pairwise similarities between topic embeddings,
and consider these similarities when computing the similarity between
two projected topic proportion vectors. Two similar documents will
then still receive a high similarity score.

\section{Experimental Results}

To investigate the quality of document representation of our TopicVec
model, we compared its performance against eight topic modeling or
document representation methods in two document classification tasks.
Moreover, to show the topic coherence of TopicVec on a single document,
we present the top words in top topics learned on a news article.

\subsection{Document Classification Evaluation}

\subsubsection{Experimental Setup}

\textbf{Compared Methods} \enskip{}Two setups of TopicVec were evaluated:
\begin{itemize}[topsep=4pt,itemsep=-0.6ex]
\item \textbf{TopicVec}: the topic proportions learned by TopicVec;
\item \textbf{TV+MeanWV}: the topic proportions, concatenated with the mean
word embedding of the document (same as the MeanWV below).\vspace{-2bp}

\end{itemize}
We compare the performance of our methods against eight methods, including
three topic modeling methods, three continuous document representation
methods, and the conventional bag-of-words (\textbf{BOW}) method.
The count vector of BOW is unweighted.

The topic modeling methods include:
\begin{itemize}[topsep=4pt,itemsep=-0.6ex]
\item \textbf{LDA}: the vanilla LDA \cite{lda} in the gensim library\footnote{https://radimrehurek.com/gensim/models/ldamodel.html};
\item \textbf{sLDA}: Supervised Topic Model\footnote{http://www.cs.cmu.edu/\textasciitilde{}chongw/slda/}
\cite{slda}, which improves the predictive performance of LDA by
modeling class labels;
\item \textbf{LFTM}: Latent Feature Topic Modeling\footnote{https://github.com/datquocnguyen/LFTM/}
\cite{lftm}.\vspace{-2bp}

\end{itemize}
The document-topic proportions of topic modeling methods were used
as their document representation. 

The document representation methods are:
\begin{itemize}[topsep=4pt,itemsep=-0.6ex]
\item \textbf{Doc2Vec}: Paragraph Vector \cite{doc2vec} in the gensim
library\footnote{https://radimrehurek.com/gensim/models/doc2vec.html}.
\item \textbf{TWE}: Topical Word Embedding\footnote{https://github.com/largelymfs/topical\_word\_embeddings/}
\cite{liu}, which represents a document by concatenating average
topic embedding and average word embedding, similar to our TV+MeanWV;
\item \textbf{GaussianLDA}: Gaussian LDA\footnote{https://github.com/rajarshd/Gaussian\_LDA}
\cite{gaussianLDA}, which assumes that words in a topic are random
samples from a multivariate Gaussian distribution with the mean as
the topic embedding. Similar to TopicVec, we derived the posterior
topic proportions as the features of each document;
\item \textbf{MeanWV}: The mean word embedding of the document.\vspace{-2bp}

\end{itemize}
\textbf{Datasets }\enskip{}We used two standard document classification
corpora: the 20 Newsgroups\footnote{http://qwone.com/\textasciitilde{}jason/20Newsgroups/}
and the ApteMod version of the Reuters-21578 corpus\footnote{http://www.nltk.org/book/ch02.html}.
The two corpora are referred to as the \textbf{20News} and \textbf{Reuters}
in the following.

20News contains about 20,000 newsgroup documents evenly partitioned
into 20 different categories. Reuters contains 10,788 documents, where
each document is assigned to one or more categories. For the evaluation
of document classification, documents appearing in two or more categories
were removed. The numbers of documents in the categories of Reuters
are highly imbalanced, and we only selected the largest 10 categories,
leaving us with 8,025 documents in total.

The same preprocessing steps were applied to all methods: words were
lowercased; stop words and words out of the word embedding vocabulary
(which means that they are extremely rare) were removed.\vspace{3bp}

\noindent \textbf{Experimental Settings}\enskip{} TopicVec used the
word embeddings trained using PSDVec on a March 2015 Wikipedia snapshot.
It contains the most frequent 180,000 words. The dimensionality of
word embeddings and topic embeddings was 500. The hyperparameters
were $\boldsymbol{\alpha}=(0.1,\cdots,0.1),\gamma=7$. For 20news
and Reuters, we specified 15 and 12 topics in each category on the
training set, respectively. The first topic in each category was always
set to null. The learned topic embeddings were combined to form the
whole topic set, where redundant null topics in different categories
were removed, leaving us with 281 topics for 20News and 111 topics
for Reuters. The initial learning rate was set to 0.1. After 100 GEM
iterations on each dataset, the topic embeddings were obtained. Then
the posterior document-topic distributions of the test sets were derived
by performing one E-step given the topic embeddings trained on the
training set.

LFTM includes two models: LF-LDA and LF-DMM. We chose the better performing
LF-LDA to evaluate. TWE includes three models, and we chose the best
performing TWE-1 to compare.

LDA, sLDA, LFTM and TWE used the specified 50 topics on Reuters, as
this is the optimal topic number according to \cite{invest}. On the
larger 20news dataset, they used the specified 100 topics. Other hyperparameters
of all compared methods were left at their default values. 

GaussianLDA was specified 100 topics on 20news and 70 topics on Reuters.
As each sampling iteration took over 2 hours, we only had time for
100 sampling iterations.

For each method, after obtaining the document representations of the
training and test sets, we trained an $\ell$-1 regularized linear
SVM one-vs-all classifier on the training set using the scikit-learn
library\footnote{http://scikit-learn.org/stable/modules/svm.html}.
We then evaluated its predictive performance on the test set.\vspace{3bp}

\noindent \textbf{Evaluation metrics} \enskip{}Considering that the
largest few categories dominate Reuters, we adopted macro-averaged
precision, recall and F1 measures as the evaluation metrics, to avoid
the average results being dominated by the performance of the top
categories.\vspace{3bp}

\noindent \textbf{Evaluation Results} \enskip{}Table 2 presents the
performance of the different methods on the two classification tasks.
The highest scores were highlighted with boldface. It can be seen
that TV+MeanWV and TopicVec obtained the best performance on the two
tasks, respectively. With only topic proportions as features, TopicVec
performed slightly better than BOW, MeanWV and TWE, and significantly
outperformed four other methods. The number of features it used was
much lower than BOW, MeanWV and TWE (Table 3).

GaussianLDA performed considerably inferior to all other methods.
After checking the generated topic embeddings manually, we found that
the embeddings for different topics are highly similar to each other.
Hence the posterior topic proportions were almost uniform and non-discriminative.
In addition, on the two datasets, even the fastest Alias sampling
in \cite{gaussianLDA} took over 2 hours for one iteration and 10
days for the whole 100 iterations. In contrast, our method finished
the 100 EM iterations in 2 hours. 

\begin{table}
\centering{}\renewcommand{\thempfootnote}{\arabic{mpfootnote}}%
\begin{minipage}[t]{1\columnwidth}%
\begin{center}
\begin{tabular}{|c|c|c|c|c|c|c|}
\hline 
\multirow{2}{*}{} & \multicolumn{3}{c|}{{\footnotesize{}20News}} & \multicolumn{3}{c|}{{\footnotesize{}Reuters}}\tabularnewline
\cline{2-7} 
 & {\footnotesize{}Prec} & {\footnotesize{}Rec} & {\footnotesize{}F1} & {\footnotesize{}Prec} & {\footnotesize{}Rec} & {\footnotesize{}F1}\tabularnewline
\hline 
\hline 
{\footnotesize{}BOW} & {\footnotesize{}69.1} & {\footnotesize{}68.5} & {\footnotesize{}68.6} & {\footnotesize{}92.5} & {\footnotesize{}90.3} & {\footnotesize{}91.1}\tabularnewline
\hline 
{\footnotesize{}LDA} & {\footnotesize{}61.9} & {\footnotesize{}61.4} & {\footnotesize{}60.3} & {\footnotesize{}76.1} & {\footnotesize{}74.3} & {\footnotesize{}74.8}\tabularnewline
\hline 
{\footnotesize{}sLDA} & {\footnotesize{}61.4} & {\footnotesize{}60.9} & {\footnotesize{}60.9} & {\footnotesize{}88.3} & {\footnotesize{}83.3} & {\footnotesize{}85.1}\tabularnewline
\hline 
{\footnotesize{}LFTM} & {\footnotesize{}63.5} & {\footnotesize{}64.8} & {\footnotesize{}63.7} & {\footnotesize{}84.6} & {\footnotesize{}86.3} & {\footnotesize{}84.9}\tabularnewline
\hline 
{\footnotesize{}MeanWV} & {\footnotesize{}70.4} & {\footnotesize{}70.3} & {\footnotesize{}70.1} & {\footnotesize{}92.0} & {\footnotesize{}89.6} & {\footnotesize{}90.5}\tabularnewline
\hline 
{\footnotesize{}Doc2Vec} & {\footnotesize{}56.3} & {\footnotesize{}56.6} & {\footnotesize{}55.4} & {\footnotesize{}84.4} & {\footnotesize{}50.0} & {\footnotesize{}58.5}\tabularnewline
\hline 
{\footnotesize{}TWE} & {\footnotesize{}69.5} & {\footnotesize{}69.3} & {\footnotesize{}68.8} & {\footnotesize{}91.0} & {\footnotesize{}89.1} & {\footnotesize{}89.9}\tabularnewline
\hline 
{\footnotesize{}GaussianLDA} & {\footnotesize{}30.9} & {\footnotesize{}26.5} & {\footnotesize{}22.7} & {\footnotesize{}46.2} & {\footnotesize{}31.5} & {\footnotesize{}35.3}\tabularnewline
\hline 
{\footnotesize{}TopicVec} & {\footnotesize{}71.3} & {\footnotesize{}71.3} & {\footnotesize{}71.2} & \textbf{\footnotesize{}92.5} & \textbf{\footnotesize{}92.1} & \textbf{\footnotesize{}92.2}\tabularnewline
\hline 
{\footnotesize{}TV+MeanWV} & \textbf{\footnotesize{}71.8} & \textbf{\footnotesize{}71.5} & \textbf{\footnotesize{}71.6} & {\footnotesize{}92.2} & {\footnotesize{}91.6} & {\footnotesize{}91.6}\tabularnewline
\hline 
\end{tabular}
\par\end{center}%
\end{minipage}\caption{Performance on multi-class text classification. Best score is in boldface.}
\end{table}

\begin{table}
\begin{centering}
\setlength\tabcolsep{3pt}%
\begin{tabular}{|c|c|c|c|c|c|}
\hline 
{\footnotesize{}Avg. Features} & {\footnotesize{}BOW} & {\footnotesize{}MeanWV} & {\footnotesize{}TWE} & {\footnotesize{}TopicVec} & {\footnotesize{}TV+MeanWV}\tabularnewline
\hline 
\hline 
{\footnotesize{}20News} & {\footnotesize{}50381} & {\footnotesize{}500} & {\footnotesize{}800} & {\footnotesize{}281} & {\footnotesize{}781}\tabularnewline
\hline 
{\footnotesize{}Reuters} & {\footnotesize{}17989} & {\footnotesize{}500} & {\footnotesize{}800} & {\footnotesize{}111} & {\footnotesize{}611}\tabularnewline
\hline 
\end{tabular}\caption{Number of features of the five best performing methods.}

\par\end{centering}

\end{table}

\subsection{Qualitative Assessment of Topics Derived from a Single Document}

Topic models need a large set of documents to extract coherent topics.
Hence, methods depending on topic models, such as TWE, are subject
to this limitation. In contrast, TopicVec can extract coherent topics
and obtain document representations even when only one document is
provided as input.

To illustrate this feature, we ran TopicVec on a New York Times news
article about a pharmaceutical company acquisition\footnote{http://www.nytimes.com/2015/09/21/business/a-huge-overnight-increase-in-a-drugs-price-raises-protests.html},
and obtained 20 topics. 

\begin{figure}
\begin{raggedright}
\hspace*{-0.3cm}\includegraphics[scale=0.24]{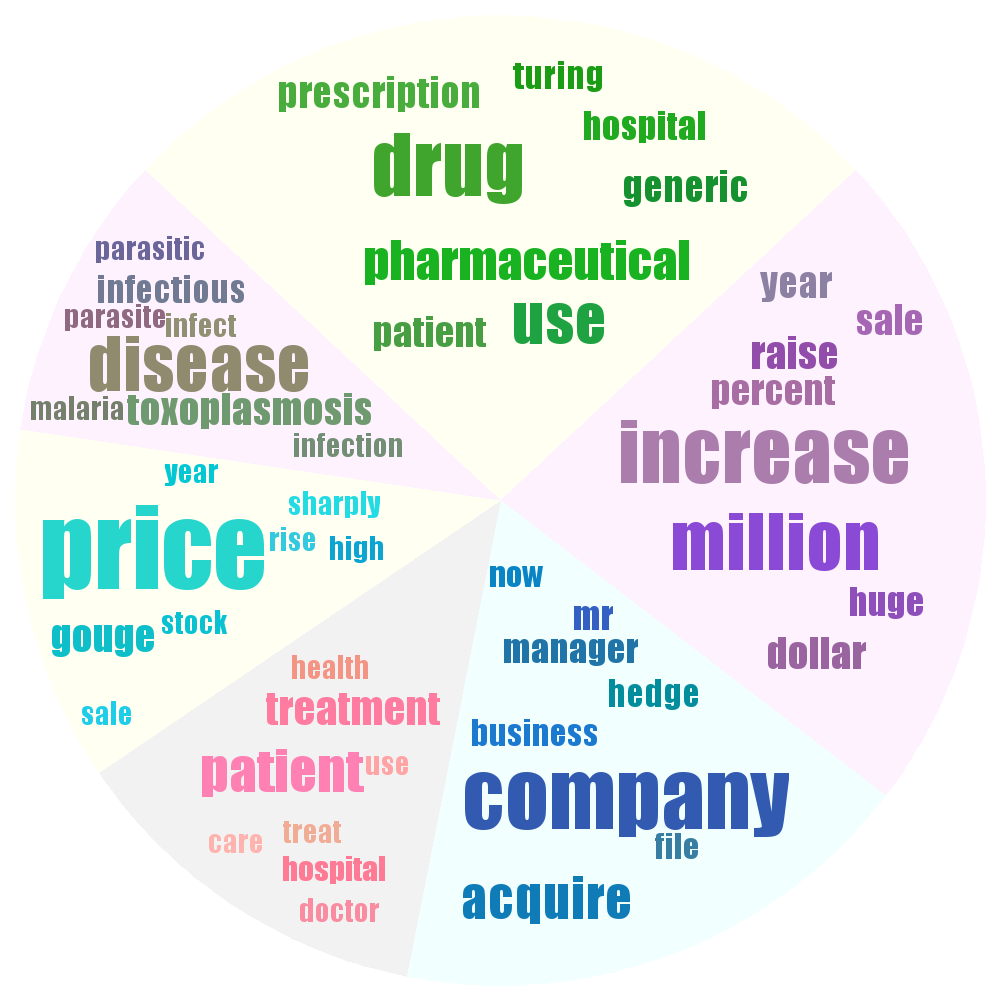}
\par\end{raggedright}

\caption{\label{fig:Topic-Cloud}Topic Cloud of the pharmaceutical company
acquisition news.}

\end{figure}

Figure \ref{fig:Topic-Cloud} presents the most relevant words in
the top-6 topics as a \emph{topic cloud}. We first calculated the
relevance between a word and a topic as the frequency-weighted cosine
similarity of their embeddings. Then the most relevant words were
selected to represent each topic. The sizes of the topic slices are
proportional to the topic proportions, and the font sizes of individual
words are proportional to their relevance to the topics. Among these
top-6 topics, the largest and smallest topic proportions are 26.7\%
and 9.9\%, respectively.

As shown in Figure \ref{fig:Topic-Cloud}, words in obtained topics
were generally coherent, although the topics were only derived from
a single document. The reason is that TopicVec takes advantage of
the rich semantic information encoded in word embeddings, which were
pretrained on a large corpus. 

The topic coherence suggests that the derived topic embeddings were
approximately the semantic centroids of the document. This capacity
may aid applications such as document retrieval, where a ``compressed
representation'' of the query document is helpful.

\section{Conclusions and Future Work}

In this paper, we proposed TopicVec, a generative model combining
word embedding and LDA, with the aim of exploiting the word collocation
patterns both at the level of the local context and the global document.
Experiments show that TopicVec can learn high-quality document representations,
even given only one document.

In our classification tasks we only explored the use of topic proportions
of a document as its representation. However, jointly representing
a document by topic proportions and topic embeddings would be more
accurate. Efficient algorithms for this task have been proposed \cite{wordmover}.

Our method has potential applications in various scenarios, such as
document retrieval, classification, clustering and summarization.

\section*{Acknowlegement}

We thank Xiuchao Sui and Linmei Hu for their help and support. We
thank the anonymous mentor for the careful proofreading. This research
is funded by the National Research Foundation, Prime Minister\textquoteright s
Office, Singapore under its IDM Futures Funding Initiative and IRC@SG
Funding Initiative administered by IDMPO. Part of the work was conceived
when Shaohua Li was visiting Tsinghua. Jun Zhu is supported by National
NSF of China (No. 61322308) and the Youngth Top-notch Talent Support
Program.\bibliographystyle{acl2016}
\bibliography{topicvec}

\begin{thebibliography}{}

\bibitem[\protect\citename{Bengio \bgroup et al.\egroup }2003]{bengio}
Yoshua Bengio, R{\'e}jean Ducharme, Pascal Vincent, and Christian Jauvin.
\newblock 2003.
\newblock A neural probabilistic language model.
\newblock {\em Journal of Machine Learning Research}, pages 1137--1155.

\bibitem[\protect\citename{Blei \bgroup et al.\egroup }2003]{lda}
David~M Blei, Andrew~Y Ng, and Michael~I Jordan.
\newblock 2003.
\newblock Latent dirichlet allocation.
\newblock {\em the Journal of machine Learning research}, 3:993--1022.

\bibitem[\protect\citename{Das \bgroup et al.\egroup }2015]{gaussianLDA}
Rajarshi Das, Manzil Zaheer, and Chris Dyer.
\newblock 2015.
\newblock Gaussian {LDA} for topic models with word embeddings.
\newblock In {\em Proceedings of the 53rd Annual Meeting of the Association for
  Computational Linguistics and the 7th International Joint Conference on
  Natural Language Processing (Volume 1: Long Papers)}, pages 795--804,
  Beijing, China, July. Association for Computational Linguistics.

\bibitem[\protect\citename{Hinton and Salakhutdinov}2009]{replicated-softmax}
Geoffrey~E Hinton and Ruslan~R Salakhutdinov.
\newblock 2009.
\newblock Replicated softmax: an undirected topic model.
\newblock In {\em Advances in neural information processing systems}, pages
  1607--1614.

\bibitem[\protect\citename{Huang \bgroup et al.\egroup }2012]{huang-global}
Eric~H Huang, Richard Socher, Christopher~D Manning, and Andrew~Y Ng.
\newblock 2012.
\newblock Improving word representations via global context and multiple word
  prototypes.
\newblock In {\em Proceedings of the 50th Annual Meeting of the Association for
  Computational Linguistics: Long Papers-Volume 1}, pages 873--882. Association
  for Computational Linguistics.

\bibitem[\protect\citename{Kusner \bgroup et al.\egroup }2015]{wordmover}
Matt Kusner, Yu~Sun, Nicholas Kolkin, and Kilian~Q. Weinberger.
\newblock 2015.
\newblock From word embeddings to document distances.
\newblock In David Blei and Francis Bach, editors, {\em Proceedings of the 32nd
  International Conference on Machine Learning (ICML-15)}, pages 957--966. JMLR
  Workshop and Conference Proceedings.

\bibitem[\protect\citename{Larochelle and Lauly}2012]{docNADE}
Hugo Larochelle and Stanislas Lauly.
\newblock 2012.
\newblock A neural autoregressive topic model.
\newblock In {\em Advances in Neural Information Processing Systems}, pages
  2708--2716.

\bibitem[\protect\citename{Le and Mikolov}2014]{doc2vec}
Quoc Le and Tomas Mikolov.
\newblock 2014.
\newblock Distributed representations of sentences and documents.
\newblock In {\em Proceedings of the 31st International Conference on Machine
  Learning (ICML-14)}, pages 1188--1196.

\bibitem[\protect\citename{Levy \bgroup et al.\egroup }2015]{levy}
Omer Levy, Yoav Goldberg, and Ido Dagan.
\newblock 2015.
\newblock Improving distributional similarity with lessons learned from word
  embeddings.
\newblock {\em Transactions of the Association for Computational Linguistics},
  3:211--225.

\bibitem[\protect\citename{Li \bgroup et al.\egroup }2015]{psdvec}
Shaohua Li, Jun Zhu, and Chunyan Miao.
\newblock 2015.
\newblock A generative word embedding model and its low rank positive
  semidefinite solution.
\newblock In {\em Proceedings of the 2015 Conference on Empirical Methods in
  Natural Language Processing}, pages 1599--1609, Lisbon, Portugal, September.
  Association for Computational Linguistics.

\bibitem[\protect\citename{Li \bgroup et al.\egroup }2016]{psdvec-osp}
Shaohua Li, Jun Zhu, and Chunyan Miao.
\newblock 2016.
\newblock {PSDVec}: a toolbox for incremental and scalable word embedding.
\newblock {\em To appear in Neurocomputing}.

\bibitem[\protect\citename{Liu \bgroup et al.\egroup }2015]{liu}
Yang Liu, Zhiyuan Liu, Tat-Seng Chua, and Maosong Sun.
\newblock 2015.
\newblock Topical word embeddings.
\newblock In {\em AAAI}, pages 2418--2424.

\bibitem[\protect\citename{Lu \bgroup et al.\egroup }2011]{invest}
Yue Lu, Qiaozhu Mei, and ChengXiang Zhai.
\newblock 2011.
\newblock Investigating task performance of probabilistic topic models: an
  empirical study of {PLSA} and {LDA}.
\newblock {\em Information Retrieval}, 14(2):178--203.

\bibitem[\protect\citename{McAuliffe and Blei}2008]{slda}
Jon~D McAuliffe and David~M Blei.
\newblock 2008.
\newblock Supervised topic models.
\newblock In {\em Advances in neural information processing systems}, pages
  121--128.

\bibitem[\protect\citename{Mikolov \bgroup et al.\egroup }2013]{word2vec}
Tomas Mikolov, Ilya Sutskever, Kai Chen, Greg~S Corrado, and Jeff Dean.
\newblock 2013.
\newblock Distributed representations of words and phrases and their
  compositionality.
\newblock In {\em Proceedings of NIPS 2013}, pages 3111--3119.

\bibitem[\protect\citename{Nguyen \bgroup et al.\egroup }2015]{lftm}
Dat~Quoc Nguyen, Richard Billingsley, Lan Du, and Mark Johnson.
\newblock 2015.
\newblock Improving topic models with latent feature word representations.
\newblock {\em Transactions of the Association for Computational Linguistics},
  3:299--313.

\bibitem[\protect\citename{Pennington \bgroup et al.\egroup }2014]{glove}
Jeffrey Pennington, Richard Socher, and Christopher~D Manning.
\newblock 2014.
\newblock {GloVe}: Global vectors for word representation.
\newblock {\em Proceedings of the Empiricial Methods in Natural Language
  Processing (EMNLP 2014)}, 12.

\end{thebibliography}
\newpage{}\onecolumn\begin{appendices}

\section{Derivation of $\mathcal{L}(q,\boldsymbol{T})$\label{sec:Derivation-of-L}}

The variational distribution is defined as:

\begin{align}
 & q(\boldsymbol{Z},\boldsymbol{\phi};\boldsymbol{\pi},\boldsymbol{\theta})=q(\boldsymbol{\phi};\boldsymbol{\theta})q(\boldsymbol{Z};\boldsymbol{\pi})\nonumber \\
= & \prod_{i=1}^{M}\left\{ \textrm{Dir}(\boldsymbol{\phi}_{i};\boldsymbol{\theta}_{i})\prod_{j=1}^{L_{i}}\textrm{Cat}(z_{ij};\boldsymbol{\pi}_{ij})\right\} .\label{eq:vardist-1}
\end{align}

Taking the logarithm of both sides of \eqref{eq:vardist-1}, we obtain
\begin{align*}
 & \log q(\boldsymbol{Z},\boldsymbol{\phi};\boldsymbol{\pi},\boldsymbol{\theta})\\
= & \sum_{i=1}^{M}\Biggl\{\log\Gamma(\theta_{i0})-\sum_{k=1}^{K}\log\Gamma(\theta_{ik})\\
 & +\sum_{k=1}^{K}(\theta_{ik}-1)\log\phi_{ik}+\sum_{j,k=1}^{L_{i},K}\delta(z_{ij}=k)\log\pi_{ij}^{k}\Biggr\},
\end{align*}
where $\theta_{i0}=\sum_{k=1}^{K}\theta_{ik}$, $\pi_{ij}^{k}$ is
the $k$-th component of $\boldsymbol{\pi}_{ij}$.

Let $\psi(\cdot)$ denote the digamma function: $\psi(x)=\frac{d}{dx}\ln{\Gamma(x)}=\frac{\Gamma'(x)}{\Gamma(x)}.$
It follows that 
\begin{align}
 & \mathcal{H}(q)\nonumber \\
= & -E_{q}[\log q(\boldsymbol{Z},\boldsymbol{\phi};\boldsymbol{\pi},\boldsymbol{\theta})]\nonumber \\
= & \sum_{i=1}^{M}\Biggl\{\sum_{k=1}^{K}\log\Gamma(\theta_{ik})-\log\Gamma(\theta_{i0})-\sum_{k=1}^{K}(\theta_{ik}-1)\psi(\theta_{ik})\nonumber \\
 & +(\theta_{i0}-K)\psi(\theta_{i0})-\sum_{j,k=1}^{L_{i},K}\pi_{ij}^{k}\log\pi_{ij}^{k}\Biggr\}.\label{eq:entropyQ-1}
\end{align}

Plugging $q$ into $\mathcal{L}(q,\boldsymbol{T})$, we have
\begin{align}
 & \mathcal{L}(q,\boldsymbol{T})\nonumber \\
= & \mathcal{H}(q)+E_{q}\left[\log p(\boldsymbol{Z},\boldsymbol{\phi}|\boldsymbol{T})\right]\nonumber \\
= & \mathcal{H}(q)+C_{0}-\log\mathcal{Z}(\boldsymbol{H},\boldsymbol{\mu})-\Vert\boldsymbol{A}\Vert_{f(\boldsymbol{H})}^{2}-\sum_{i=1}^{W}\mu_{i}\Vert\boldsymbol{v}_{s_{i}}\Vert^{2}\nonumber \\
 & +\sum_{i=1}^{M}\vast\{\sum_{k=1}^{K}\left(E_{q(\boldsymbol{Z}_{i}|\boldsymbol{\pi}_{i})}[m_{ik}]+\alpha_{k}-1\right)\cdot E_{q(\phi_{ik}|\boldsymbol{\theta}_{i})}[\log\phi_{ik}]\nonumber \\
 & +\sum_{j=1}^{L_{i}}\Bigg(\boldsymbol{v}_{w_{ij}}^{\T}\Bigl(\sum_{k=j-c}^{j-1}\boldsymbol{v}_{w_{ik}}+E_{q(z_{ij}|\boldsymbol{\pi}_{ij})}[\boldsymbol{t}_{z_{ij}}]\Bigr)+\sum_{k=j-c}^{j-1}a_{w_{ik}w_{ij}}+E_{q(z_{ij}|\boldsymbol{\pi}_{ij})}[r_{i,z_{ij}}]\Bigg)\vast\}\nonumber \\
= & C_{1}+\mathcal{H}(q)+\sum_{i=1}^{M}\Biggl\{\sum_{k=1}^{K}\Bigl(\sum_{j=1}^{L_{i}}\pi_{ij}^{k}+\alpha_{k}-1\Bigr)\Bigl(\psi(\theta_{ik})-\psi(\theta_{i0})\Bigr)+\sum_{j=1}^{L_{i}}\Bigl(\boldsymbol{v}_{w_{ij}}^{\T}\boldsymbol{T}_{i}\boldsymbol{\pi}_{ij}+\boldsymbol{r}_{i}^{\T}\boldsymbol{\pi}_{ij}\Bigr)\Biggr\}\nonumber \\
= & C_{1}+\mathcal{H}(q)+\sum_{i=1}^{M}\Biggl\{\sum_{k=1}^{K}\Bigl(\sum_{j=1}^{L_{i}}\pi_{ij}^{k}+\alpha_{k}-1\Bigr)\Bigl(\psi(\theta_{ik})-\psi(\theta_{i0})\Bigr)\nonumber \\
 & +\textrm{Tr}(\boldsymbol{T}_{i}^{\T}\sum_{j=1}^{L_{i}}\boldsymbol{v}_{w_{ij}}\boldsymbol{\pi}_{ij}^{\T})+\boldsymbol{r}_{i}^{\T}\sum_{j=1}^{L_{i}}\boldsymbol{\pi}_{ij}\Biggr\},\label{eq:emobj-1}
\end{align}

\section{Derivation of the E-Step\label{sec:E-step}}

The learning objective is:

\begin{align}
 & \mathcal{L}(q,\boldsymbol{T})\nonumber \\
= & \sum_{i=1}^{M}\Biggl\{\sum_{k=1}^{K}\Bigl(\sum_{j=1}^{L_{i}}\pi_{ij}^{k}+\alpha_{k}-1\Bigr)\Bigl(\psi(\theta_{ik})-\psi(\theta_{i0})\Bigr)\nonumber \\
 & +\textrm{Tr}(\boldsymbol{T}_{i}^{\T}\sum_{j=1}^{L_{i}}\boldsymbol{v}_{w_{ij}}\boldsymbol{\pi}_{ij}^{\T})+\boldsymbol{r}_{i}^{\T}\sum_{j=1}^{L_{i}}\boldsymbol{\pi}_{ij}\Biggr\}+\mathcal{H}(q)+C_{1},\label{eq:emobj-1}
\end{align}

\eqref{eq:emobj-1} can be expressed as 
\begin{align}
 & \mathcal{L}(q,\boldsymbol{T}^{(l-1)})\nonumber \\
= & \sum_{i=1}^{M}\Biggl\{\sum_{k=1}^{K}\log\Gamma(\theta_{ik})-\log\Gamma(\theta_{i0})-\sum_{k=1}^{K}(\theta_{ik}-1)\psi(\theta_{ik})+(\theta_{i0}-K)\psi(\theta_{i0})-\sum_{j,k=1}^{L_{i},K}\pi_{ij}^{k}\log\pi_{ij}^{k}\nonumber \\
 & +\sum_{k=1}^{K}\Bigl(\sum_{j=1}^{L_{i}}\pi_{ij}^{k}+\alpha_{k}-1\Bigr)\Bigl(\psi(\theta_{ik})-\psi(\theta_{i0})\Bigr)+\sum_{j=1}^{L_{i}}\Bigl(\boldsymbol{v}_{w_{ij}}^{\T}\boldsymbol{T}_{i}\boldsymbol{\pi}_{ij}+\boldsymbol{r}_{i}^{\T}\boldsymbol{\pi}_{ij}\Bigr)\Biggr\}+C_{1}.\label{eq:eobj-1}
\end{align}

We first maximize \eqref{eq:eobj-1} w.r.t. $\pi_{ij}^{k}$, the probability
that the $j$-th word in the $i$-th document takes the $k$-th latent
topic. Note that this optimization is subject to the normalization
constraint that $\sum_{k=1}^{K}\pi_{ij}^{k}=1$.

We isolate terms containing $\boldsymbol{\pi}_{ij}$, and form a Lagrange
function by incorporating the normalization constraint:
\begin{equation}
\Lambda(\boldsymbol{\pi}_{ij})=-\sum_{k=1}^{K}\pi_{ij}^{k}\log\pi_{ij}^{k}+\sum_{k=1}^{K}\Bigl(\psi(\theta_{ik})-\psi(\theta_{i0})\Bigr)\pi_{ij}^{k}+\boldsymbol{v}_{w_{ij}}^{\T}\boldsymbol{T}_{i}\boldsymbol{\pi}_{ij}+\boldsymbol{r}_{i}^{\T}\boldsymbol{\pi}_{ij}+\lambda_{ij}(\sum_{k=1}^{K}\pi_{ij}^{k}-1).\label{eq:LagrangePi-1}
\end{equation}

Taking the derivative w.r.t. $\pi_{ij}^{k}$, we obtain 
\begin{equation}
\frac{\partial\Lambda(\boldsymbol{\pi}_{ij})}{\partial\pi_{ij}^{k}}=-1-\log\pi_{ij}^{k}+\psi(\theta_{ik})-\psi(\theta_{i0})+\boldsymbol{v}_{w_{ij}}^{\T}\boldsymbol{t}_{ik}+r_{ik}+\lambda_{ij}.\label{eq:dLagdPi-1}
\end{equation}

Setting this derivative to $0$ yields the maximizing value of $\pi_{ij}^{k}$:
\begin{equation}
\pi_{ij}^{k}\propto\exp\{\psi(\theta_{ik})+\boldsymbol{v}_{w_{ij}}^{\T}\boldsymbol{t}_{ik}+r_{ik}\}.\label{eq:solPi-1}
\end{equation}

Next, we maximize \eqref{eq:eobj-1} w.r.t. $\theta_{ik}$, the $k$-th
component of the posterior Dirichlet parameter:
\begin{align}
 & \frac{\partial\mathcal{L}(q,\boldsymbol{T}^{(l-1)})}{\partial\theta_{ik}}\nonumber \\
= & \frac{\partial}{\partial\theta_{ik}}\biggl\{\log\Gamma(\theta_{ik})-\log\Gamma(\theta_{i0})+\Bigl(\sum_{j=1}^{L_{i}}\pi_{ij}^{k}+\alpha_{k}-\theta_{ik}\Bigr)\psi(\theta_{ik})-\Bigl(L_{i}+\sum_{k}\alpha_{k}-\theta_{i0}\Bigr)\psi(\theta_{i0})\biggr\}\nonumber \\
= & \Bigl(\sum_{j=1}^{L_{i}}\pi_{ij}^{k}+\alpha_{k}-\theta_{ik}\Bigr)\psi'(\theta_{ik})-\Bigl(L_{i}+\sum_{k}\alpha_{k}-\theta_{i0}\Bigr)\psi'(\theta_{i0}),\label{eq:dLdTheta-1}
\end{align}
where $\psi'(\cdot)$ is the derivative of the digamma function $\psi(\cdot)$,
commonly referred to as the \textit{trigamma function}.

Setting \eqref{eq:dLdTheta-1} to $0$ yields a maximum at
\begin{equation}
\theta_{ik}=\sum_{j=1}^{L_{i}}\pi_{ij}^{k}+\alpha_{k}.\label{eq:solTheta-1}
\end{equation}

Note this solution depends on the values of $\pi_{ij}^{k}$, which
in turn depends on $\theta_{ik}$ in \eqref{eq:solPi-1}. Then we
have to alternate between \eqref{eq:solPi-1} and \eqref{eq:solTheta-1}
until convergence.

\section{Derivative of $\mathcal{L}(q^{(l)},\boldsymbol{T})$ w.r.t. $\boldsymbol{T}_{i}$:\label{sec:dLdT}}

\begin{align}
 & \frac{\partial\mathcal{L}(q^{(l)},\boldsymbol{T})}{\partial\boldsymbol{T}_{i}}\nonumber \\
= & \frac{\partial\sum_{j=1}^{L_{i}}\left(\boldsymbol{v}_{w_{ij}}^{\T}\boldsymbol{T}_{i}\boldsymbol{\pi}_{ij}+\boldsymbol{\pi}_{ij}^{\T}\boldsymbol{r}_{i}\right)}{\partial\boldsymbol{T}_{i}}\nonumber \\
= & \frac{\partial}{\partial\boldsymbol{T}_{i}}\textrm{Tr}(\boldsymbol{T}_{i}\sum_{j=1}^{L_{i}}\boldsymbol{\pi}_{ij}\boldsymbol{v}_{w_{ij}}^{\T})+(\sum_{j=1}^{L_{i}}\boldsymbol{\pi}_{ij})^{\T}\frac{\partial\boldsymbol{r}_{i}}{\partial\boldsymbol{T}_{i}}\nonumber \\
= & \sum_{j=1}^{L_{i}}\boldsymbol{v}_{w_{ij}}\boldsymbol{\pi}_{ij}^{\T}+(\sum_{j=1}^{L_{i}}\boldsymbol{\pi}_{ij})^{\T}\frac{\partial\boldsymbol{r}_{i}}{\partial\boldsymbol{T}_{i}}\nonumber \\
= & \sum_{j=1}^{L_{i}}\boldsymbol{v}_{w_{ij}}\boldsymbol{\pi}_{ij}^{\T}+\sum_{k=1}^{K}\bar{m}_{ik}\frac{\partial r_{ik}}{\partial\boldsymbol{T}_{i}},\label{eq:dLdT-1}
\end{align}
where $\bar{m}_{ik}=\sum_{j=1}^{L_{i}}\pi_{ij}^{k}=E[m_{ik}]$, the
sum of the variational probabilities of each word being assigned to
the $k$-th topic in the $i$-th document.

\end{appendices}
\end{document}